\documentclass[11pt]{article}

\usepackage[final]{acl}
\usepackage{multirow}

\usepackage{times}
\usepackage{latexsym}

\usepackage[T1]{fontenc}
\usepackage[utf8]{inputenc}

\usepackage{microtype}

\usepackage{inconsolata}

\usepackage{graphicx}

\usepackage{float}

\usepackage{booktabs}

\title{Beyond Many-Shot Translation: Scaling In‑Context Demonstrations For Low‑Resource Machine Translation}

\author{
 \textbf{Luis Frentzen Salim\textsuperscript{1,2}},
 \textbf{Esteban Carlin\textsuperscript{2,3}},
 \textbf{Alexandre Morinvil\textsuperscript{2}},
 \textbf{Xi Ai\textsuperscript{4}},
 \textbf{Lun-Wei Ku\textsuperscript{1}}
\\
\\
 \textsuperscript{1}Institute of Information Science, Academia Sinica, \\
 \textsuperscript{2}National Taiwan University of Science and Technology, \\
 \textsuperscript{3}Ecole Centrale de Marseille,
 \textsuperscript{4}National University of Singapore,
\\
 \small{
   \textbf{Correspondence:} \href{mailto:luisfrentzen@gmail.com}{luisfrentzen@gmail.com}, \href{mailto:esteban.carlin@centrale-med.fr}{esteban.carlin@centrale-med.fr}, \href{mailto:alexandre.morinvil@outlook.com}{alexandre.morinvil@outlook.com}
 }
}

\begin{document}
\maketitle
\begin{abstract}
Building machine translation (MT) systems for low-resource languages is notably difficult due to the scarcity of high-quality data. Although Large Language Models (LLMs) have improved MT system performance, adapting them to lesser-represented languages remains challenging. In-context learning (ICL) may offer novel ways to adapt LLMs for low-resource MT by conditioning models on demonstration at inference time. In this study, we explore scaling low-resource machine translation ICL beyond the few-shot setting to thousands of examples with long-context models. We scale in-context token budget to 1M tokens and compare three types of training corpora used as in-context supervision: monolingual unsupervised data, instruction-style data, and parallel data (English--target and Indonesian--target). Our experiments on Javanese and Sundanese show that gains from additional context saturate quickly and can degrade near the maximum context window, with scaling behavior strongly dependent on corpus type. Notably, some forms of monolingual supervision can be competitive with parallel data, despite the latter offering additional supervision. Overall, our results characterize the effective limits and corpus-type sensitivity of long-context ICL for low-resource MT, highlighting that larger context windows do not necessarily yield proportional quality gains.
\end{abstract}

\section{Introduction}

Machine Translation (MT) systems for low-resource languages face persistent challenges due to the scarcity of high-quality data~\cite{koehn-knowles-2017-six, maruf2023survey}. While the advancement of Large Language Models (LLMs) has improved the quality and the range of MT systems~\cite{robinson-etal-2023-chatgpt, bawden-yvon-2023-investigating}, modeling low-resource languages remains difficult. Common approaches such as architecture adaptation, transfer learning, or supervised finetuning typically rely on large amounts of high-quality supervision, which is often impractical when training data are constrained. 

LLMs offer a complementary way to improve task performance without modifying the model's architecture or parameters. By placing a few examples in context, we enable in-context learning (ICL) that could improve task performance and accuracy, so long as the model's context window allows~\cite{NEURIPS2020_1457c0d6, lyu2024paradigmshiftfuturemachine,chitale-etal-2024-empirical}. The emergence of long-context LLMs further extends this paradigm to a many-shot setting, where the context can contain hundreds to thousands of examples. Prior work~\cite {agarwal2024manyshotincontextlearning} demonstrates that scaling few-shot prompting to hundreds of examples can achieve performance comparable to finetuning on some tasks, including machine translation. Furthermore, techniques like prompt prefix caching can be used to scale long-context systems, especially for static in-context examples, thereby mitigating inference latency.

However, several questions remain open, especially for low-resource translation. For example, it is unclear whether performance will continue to improve as the number of in-context examples grows. Previous experiments were testing only up to hundreds of examples within the commonly used 128K token window. Furthermore, these studies usually draw examples and evaluation samples from the same data distribution, which provides hidden benefits for tasks such as machine translation and may not be practical in real-world scenarios where data availability is a major problem. 

In this work, we hypothesize that both the amount and the source of ICL examples influence translation performance, and that there is a limit beyond which additional context yields diminishing or negative returns. We also aim to test how ICL would scale with different types and distributions of in-context demonstrations. To test this hypothesis, we examine how translation performance scales with the number of in-context examples, using demonstrations sourced from three types of datasets: monolingual corpora, instruction-style data, and parallel corpora. In this study, we focus on Javanese and Sundanese as low-resource target languages. Our contribution is three-fold:
\begin{enumerate}
    \item \textbf{Many-shot low-resource MT scaling characterization up to 1M tokens.} We characterize many-shot in-context learning for low-resource language translation to the extent of 1M tokens. We observe that, in some cases, the effective context size for improving translation quality is often smaller than the maximum context limit.
    \item \textbf{Data-type sensitivity.} We study how different ICL prompting corpora can also affect the performance yield and the efficiency of long in-context learning for machine translation.
    \item \textbf{Practical implications for low-resource prompting.} We explore novel ways and directions to help low-resource language translation via ICL, considering that the amount of high-quality data of certain languages restricts the possibility of data-heavy alternatives. Additionally, we release the constructed translated parallel corpora for English-Indonesian-Javanese and English-Indonesian-Sundanese triplets.
\end{enumerate}

While our study focuses on translation, similar dynamics may extend to other tasks, reflecting broader trends in long-context modeling and the scaling behavior of in-context learning~\cite{dong-etal-2024-survey,agarwal2024manyshotincontextlearning}.

\begin{figure}[t]
  \centering
  \setlength{\fboxsep}{8pt}
  \fbox{%
  \begin{minipage}{0.9\columnwidth}
    \ttfamily\small
    \textbf{Prompt}:\\
    You are a helpful translation assistant. Your current task is to translate texts as accurately as possible.\\
    \\
    \{Shot Samples\}\\
    \\
    Translate the following sentence from \{Source Lang Label\} to \{Target Lang Label\}.\\
    \{Source Lang Label\} sentence:\\
    \{Source Lang Sentence\}\\
  \end{minipage}
  }
  \caption{Main prompt template used for evaluation. Different type of data used in shot samples will have their own formats, which is described in later sections.}
  \label{fig:prompttemplate}
\end{figure}

\section{Related Work}

\subsection{Low-resource Machine Translation}

Most NLP research has focused on only a small subset of the world's languages, and MT is no exception \cite{joshi-etal-2020-state, fan2020englishcentricmultilingualmachinetranslation}. 
For low-resource languages, high-quality MT remains challenging primarily due to limited parallel data \cite{haddow-etal-2022-survey}. Data collection research aims to close the gap by collecting high-quality data through data crawling and, most often, crowd sourcing \cite{cahyawijaya-etal-2025-crowdsource, lovenia-etal-2024-seacrowd}. However, data collection remained an expensive and tiring process. As a result, low-resource MT often relies on transfer from related high-resource languages and data-efficient adaptation strategies~\cite{haddow-etal-2022-survey, nguyen-chiang-2017-transfer, nllb2022}. 

\subsection{ICL for Machine Translation}

In-context learning has become a prominent paradigm for applying LLMs in MT. Early work showed that multilingual LLMs can surpass supervised baselines using only a small number of demonstrations~\cite{NEURIPS2020_1457c0d6,lin-etal-2022-shot}. Subsequent studies highlight that ICL effectiveness in MT is highly sensitive to example selection, ordering, and domain alignment~\cite{agrawal-etal-2023-context,chitale-etal-2024-empirical, zebaze2024incontextexampleselectionsimilarity}. For example, ~\citet{agrawal-etal-2023-context} shows that similarity-based retrieval with recall-oriented re-ranking yields substantial gains over random prompts, especially when in-domain examples are available. ~\citet{chitale-etal-2024-empirical} further demonstrates that MT ICL is largely example-driven, with target-side quality and spatial proximity of demonstrations playing a central role, and that related tasks with similar target distributions can suffice as in-context supervision. These findings motivate the question of whether different demonstrations (e.g., unsupervised or instruction-style corpus) lead to different scaling behavior.

\subsection{Long-Context Models}

Long context models enable processing up to millions of tokens, enabling LLMs to attend over a wide range of things at the same time~\cite{geminiteam2024gemini15unlockingmultimodal,qwen2.5-1m,ulralong2025}. However, empirical analyses suggest that models do not always exploit the full context effectively and may suffer from position-related failures, such as the lost-in-the-middle phenomenon~\cite {liu-etal-2024-lost}. Prior work, such as \citet{yuan-etal-2024-focused, agarwal2024manyshotincontextlearning}, also found that, even with hundreds of examples, the optimal result may not come from the setting with the most demonstrations. Our work studies how these long-context limitations interact with low-resource MT.

\subsection{Many-Shot In-Context Learning}

The many-shot ICL has only recently become accessible as context windows have expanded. \citet{agarwal2024manyshotincontextlearning} systematically study many-shot ICL with up to hundreds of demonstrations and report strong gains across tasks, including low-resource MT, when scaling the number of in-context examples. Their results also suggest that many-shot prompting can approach or match specialized systems on some tasks, but also reveal task-dependent behavior and non-monotonic performance trends as context grows. While the performance gain is promising, the work primarily relies on in-distribution data, which may not reflect real-world scenarios for low-resource language modeling and MT.

\begin{figure}[t]
\centering
\setlength{\fboxsep}{8pt}
\fbox{%
  \begin{minipage}{0.9\columnwidth}
  \ttfamily\small
  \textbf{Sample Format}:\\
  \{Instructions\}\\
  \{Input\}\\
  \{Output\}\\ 
  \end{minipage}
}
\caption{Sample format for each instructions sample. The Alpaca dataset contains at least an instruction, input, and output. Each sample will be concatenated and put in the shot samples section of the main prompt.}
\label{fig:instprompttemplate}
\end{figure}

\begin{figure}[t]
  \centering
  \setlength{\fboxsep}{8pt}
  \fbox{%
  \begin{minipage}{0.9\columnwidth}
    \ttfamily\small
  \textbf{Sample Format}:\\
  \{Source Lang Label\}: \{Source Lang Sentence\}\\
  \{Target Lang Label\}: \{Target Lang Sentence\}\\ 
  \end{minipage}
}
  \caption{Sample format for each parallel/supervised sample.}
  \label{fig:paraprompttemplate}
\end{figure}

\section{Methodology}

In standard MT setups, improvements typically come from three types of supervision: \textbf{(1)} Monolingual corpora for back-translation or unsupervised neural machine translation, \textbf{(2)} Instruction data for instruction finetuning or multi-task training, and \textbf{(3)} Parallel corpora for explicit translation knowledge training. This study explores whether similar benefits can be induced purely in-context instead of training and updating the parameters with the little data available for low-resource languages. We treat each corpus as an additional in-context training demonstration or shots and insert it within the model's context window, effectively running many-shot MT. The prompt template we use can be seen in Fig.~\ref{fig:prompttemplate}. This study focuses on two low-resource languages, that is \textbf{Javanese} and \textbf{Sundanese}. Both languages are great candidates for our study, as recent efforts have expanded data resources and improved baseline modeling for both languages, making systematic evaluation feasible while preserving the characteristics and challenges of low-resource machine translation. We also include parallel supervised datasets with \textbf{Indonesian} and \textbf{English} as reference languages. \textbf{Indonesian} is chosen alongside the \textbf{English} baseline, as it is typologically closer to both Javanese and Sundanese, allowing us to test whether typological closeness improves in-context learning MT into and out of the low-resource targets.

\begin{figure*}[t]
  \includegraphics[width=\textwidth]{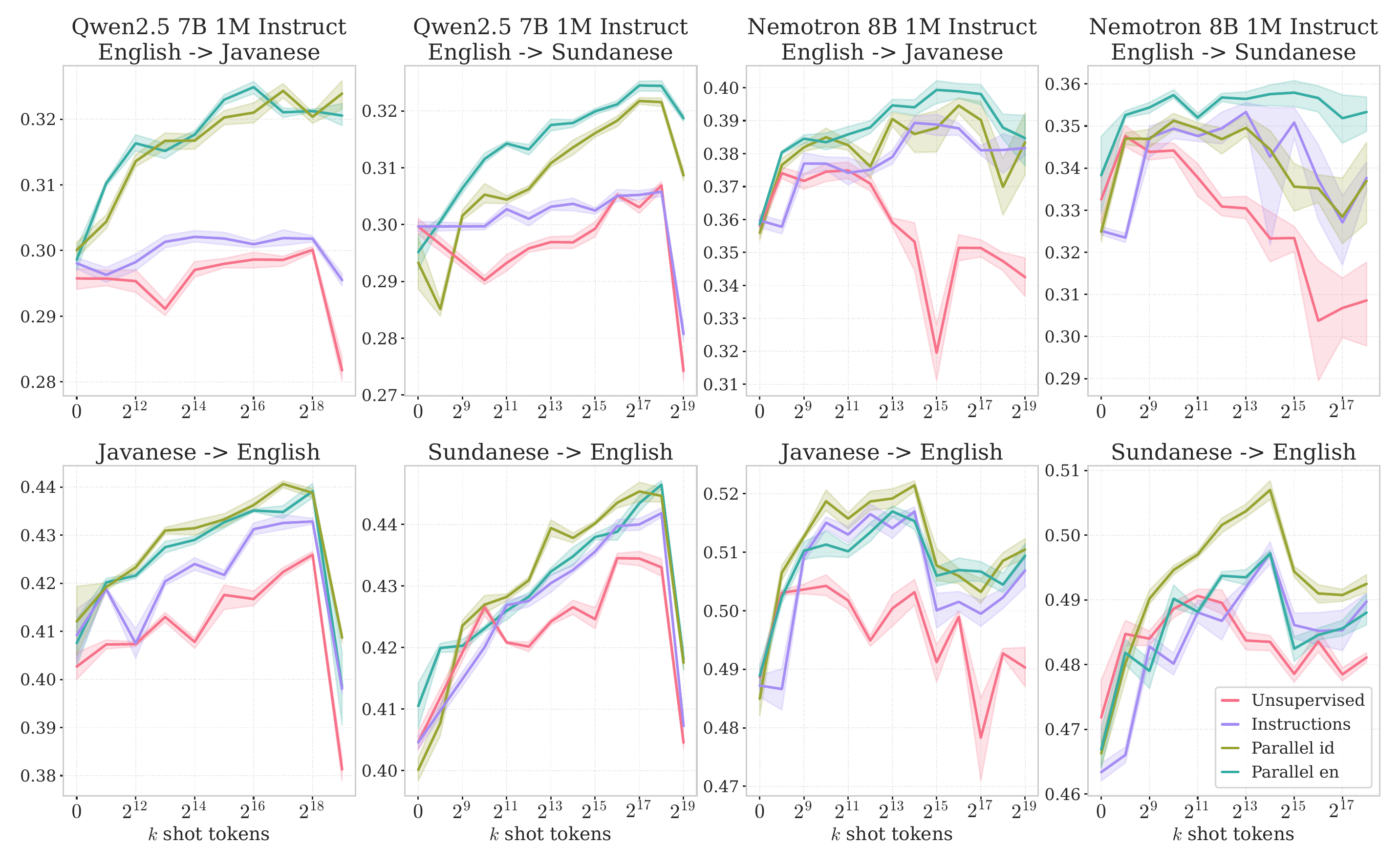}
  \caption{ChrF++ scores for two models and two language combinations. Each figure presented different lines indicating the different corpus types used as demonstrations during evaluation. We observe an initial increase in performance followed by a plateau, then a final drop. The Instruction type corpus (Purple) performs comparably to the supervised corpora despite being monolingual, showing that data presentation matters.}
  \label{fig:chrf}
\end{figure*}

\subsection{Model}

% We study two decoder-only transformers models that are \textit{Qwen2.5-7B-Instruct-1M} and \textit{Nemotron-8B-UltraLong-1M-Instruct}. Those Both are designed to process up to one million tokens in their context window \cite{qwen2.5, qwen2.5-1m, ulralong2025}. 

We study two decoder-only Transformer models, \textit{Qwen2.5-7B-Instruct-1M} and \textit{Nemotron-8B-UltraLong-1M-Instruct}. These models were selected for their ultra-long context support, enabling them to process up to one million tokens \cite{qwen2.5, qwen2.5-1m, ulralong2025}. The 8B parameter range was selected to balance computational efficiency and performance, facilitating evaluations across a wide range of configurations, with the KV cache size being the major bottleneck. Inferences were performed using vLLM~\cite{vLLM}, with a decoding temperature of $0.7$ to introduce a moderate level of stochasticity in the generated outputs.

\subsection{In-context Corpora}

We test three different types of corpora that are commonly used to train MT systems or to adapt an LLM to a new language. Each corpus is processed into a sequence of in-context examples and inserted into the prompt.

\textbf{Unsupervised Monolingual Data}. Monolingual text is commonly used for neural language models and MT systems pretraining. We use the crawled monolingual corpora provided by \cite{conneau-etal-2020-unsupervised, wenzek-etal-2020-ccnet}. The dataset was constructed from URLs and paragraphs indexed by processing the January-December 2018 Commoncrawl snapshots.
    
\textbf{Instructions Data}. We incorporate instruction-style data that is commonly used for instruction finetuning. Prior work has shown the effectiveness of using instruction or task datasets for multitask fine-tuning, which helps improve generalization even in a massively multilingual setting \cite{muennighoff-etal-2023-crosslingual}. We use the Alpaca instructions dataset, which has been translated into many languages, including Javanese and Sundanese, using Google Translate, as released in prior work~\cite{alpaca, upadhayay2024taco}. From each row of the dataset, we construct a demonstration which prompts a format shown in Fig.~\ref{fig:instprompttemplate}.

\textbf{Parallel Data}. Lastly, we use parallel corpora, which provide direct supervision by aligning a reference language with the target language. We provide two types of parallel supervision: \textbf{(1)} English--target, and \textbf{(2)} Indonesian--target. We include Indonesian as a second reference language because it is linguistically closer to the target languages than English, sharing greater structural and cultural proximity, as well as significant vocabulary overlap. 
Parallel corpora are often the hardest resource to obtain for low-resource languages. We therefore utilize the Aya collection dataset \cite{singh2024aya}, which includes text in Javanese and Sundanese. We sampled sentences which amounts to roughly 500K tokens from the Aya collection dataset. We then generated synthetic parallel sentences by translating the sampled examples from the dataset into English and Indonesian using OpenAI's GPT5 model. We translate from the target languages into the reference languages, rather than the other way around, because translating into a higher-resource language is more robust to translation errors. The parallel corpora for both Javanese and Sundanese will be released along with this paper. We then construct each demonstration using the format shown in Fig.~\ref{fig:paraprompttemplate}. More details regarding the construction of the synthetic parallel corpus can be found in Appendix~\ref{app:parallel}.

Example prompts and sample data for each corpus type can be found in Appendix~\ref{app:prompts} and Appendix~\ref{app:data}, respectively.

\begin{figure*}[t]
  \includegraphics[width=\textwidth]{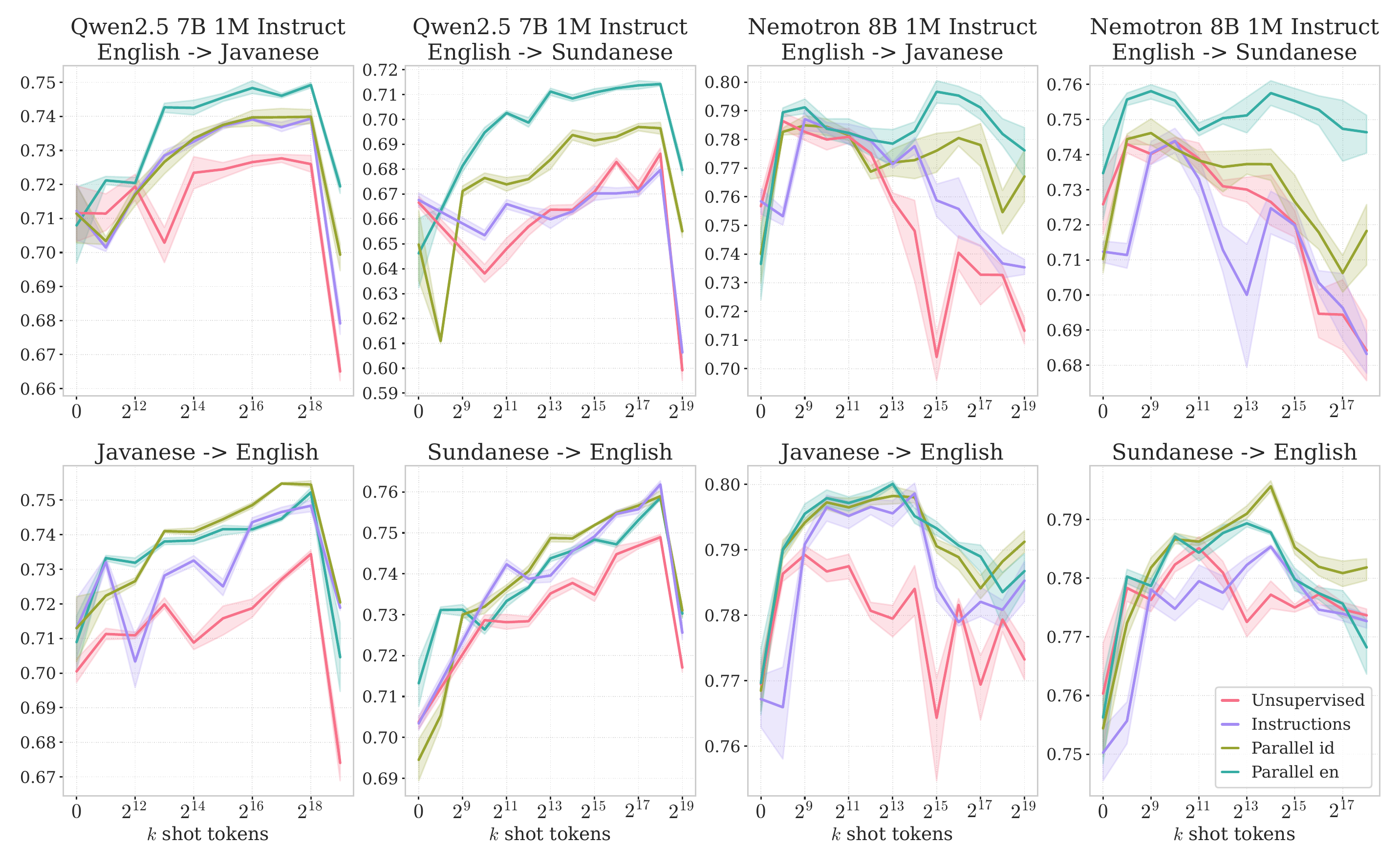}
  \caption{COMET scores for two models across two language combinations. Each subplot shows lines representing the different corpus types used as demonstrations during evaluation. We observe a pattern similar to the ChrF++ plot, an initial increase in performance, followed by a plateau, then a final drop. The relative ranking of corpus types is consistent with the ChrF++ metric.}
  \label{fig:comet}
\end{figure*}

\subsection{Evaluation}

We evaluate translation performance on the FLORES+ dataset \cite{nllb-24, goyal-etal-2022-flores, guzman-etal-2019-flores, nllb2022}. We run experiments with zero-shot and many-shot with demonstration/shot token budgets ranging from $2^{7}$ to $2^{20}$ (128 to \textasciitilde1M tokens or maximum context window), increasing by powers of 2. We will use the term \emph{demonstration tokens} or \emph{shot tokens} interchangeably to describe the example token count present in context.
We evaluate translation in both directions, from the reference languages (English, Indonesian) to the target languages (Javanese, Sundanese) and vice versa. To evaluate the translation quality, we use the COMET-22 \cite{rei-etal-2022-comet}, a neural-based evaluation metric that leverages cross-lingual embeddings to predict human quality judgments and ChrF++ metrics ($\beta=2, n_{char}=6, n_{word}=2$) \cite{popovic-2015-chrf}, which measures character and word n-gram overlap between hypothesis and reference translations. Both metrics have been shown to correlate well with human judgments \cite{kocmi-etal-2022-ms}.

For each combination of corpus context type and context size, we evaluate for five runs and report the average across runs\footnote{With the exception of some variants at the 1M token position due to compute resource constraints.}. To ensure controlled scaling of examples, we always sample the first $k$ rows of the dataset which does not exceed the $2^{n}$ context token budget. We also run a random-sampling version of the experiment, where we randomly sample $k$ rows instead of always sampling the first $k$ rows, to test the effects of recency bias and order-sensitivity on translation quality.

\subsection{Experimental Setup}

We use the vLLM generation backend for inference on NVIDIA H100 GPUs, while COMET-22 evaluation is performed on NVIDIA RTX 4090 GPUs. We additionally utilize vLLM’s prompt prefix caching to enable efficient inference in long-context settings.

\section{Results and Discussion}

\subsection{Effective Context Saturates Early}

Across both models and target languages, scaling in-context demonstrations yields an average of \textasciitilde+0.04 points in ChrF++ score and +0.05-0.06 points in COMET score. However, performance does not scale linearly with context size. It plateaus and decreases as the number of demonstrations increases, nears the limits of the model's maximum context window, as seen in both Fig.~\ref{fig:chrf} and Fig.~\ref{fig:comet}. Across most experiments, performance increases as the number of demonstration tokens grows, then quickly saturates. We compare ChrF++ scores between the first-$k$ sampling and random-$k$ sampling but find no noticeable performance difference. A paired t-test revealed no significant difference between the two conditions ($t=1.0264, p=0.3054$). More results and discussion on the random sampling experiment can be found in Appendix~\ref{app:sampling}.

Notably, performance gain starts to saturate after around $2^{14}$ and $2^{16}$ demonstration tokens, after which adding more context provides little additional benefit. The extent of performance degradation, however, varies by model and corpus type, with some configurations showing more pronounced drops than others.
% and can even degrade quality. 
This is far below the maximum 1M token limit, suggesting that the \emph{effective} context size is often smaller than the maximum supported context window. While adding more demonstration tokens naturally increases lexical/vocabulary coverage and average token frequency, which intuitively should benefit language modeling, our observation indicates diminishing returns. One plausible explanation is dispersed attention, when the average model attention for a query gets dispersed as the number of demonstrations grows, thus hurting context understanding \cite{yuan-etal-2024-focused}. This effect is naturally exacerbated in the long-context models.

\subsection{Performance Collapse Nearing Maximum Context Length}

\begin{figure}[t]
  \includegraphics[width=\columnwidth]{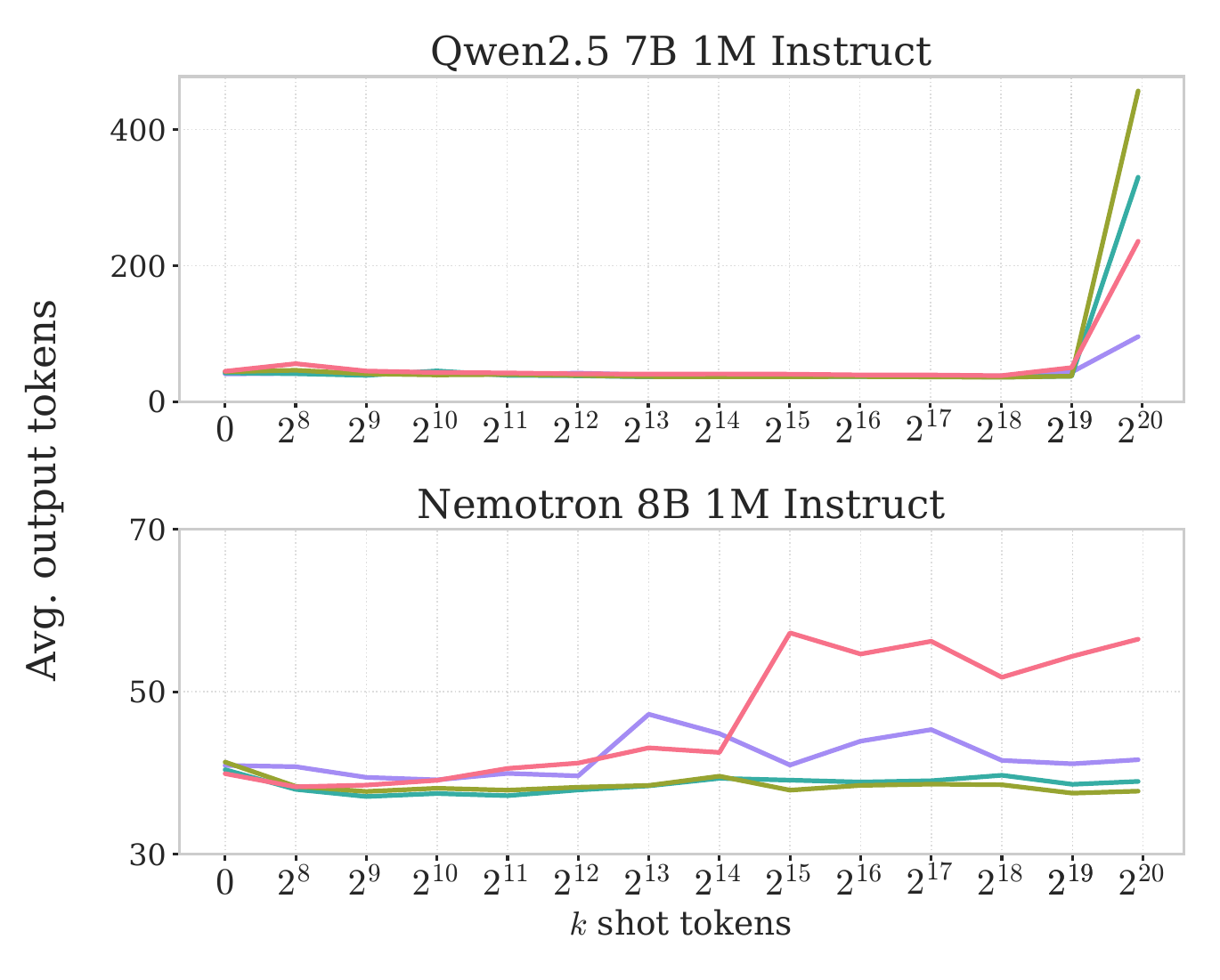}
  \caption{Sudden increase of average output token count nearing the maximum context size for the Qwen model. While the Nemotron model also increases, the increase is less pronounced than Qwen's. Output tokens for each shot token position are averaged across translation direction and target languages.}
  \label{fig:outputtoken}
\end{figure}

% We observe a pronounced performance drop roughly at 50\%--100\% of the context window, sometimes to even lower than the zero-shot performance.
We observe a pronounced performance drop around 50\%--100\% of the context window, with the severity and onset varying by model architecture and corpus type. In some configurations, particularly with Qwen on certain corpus types, performance can drop substantially, occasionally approaching or falling below zero-shot performance.
% We find that some plots resemble the downward parabola on logarithmic scale, where performance starts to decrease after peaking, and eventually drops. 
Examining the scaling curves in Fig.~\ref{fig:chrf} and Fig.~\ref{fig:comet}, we observe that performance typically peaks at intermediate context sizes (around $2^{14}$ to $2^{16}$ tokens) before declining. The decline pattern varies, some configurations show gradual degradation while others, particularly Qwen with parallel corpora, exhibit more abrupt drops at the highest context levels tested.
We also observe a noticeable increase in average output length near the maximum context limit for the Qwen model, as shown in Fig.~\ref{fig:outputtoken}, indicating the model's failure to process long-context prompts normally. We experimented with truncation to mitigate such effects, however, there was minimal change in the end result. Furthermore, this pattern is less pronounced in the Nemotron model, yet a similar drop in performance still occurs. A side observation of the output token count behavior is that different models reacted differently to different corpus types. While both parallel settings experience the highest increase of the average output token count in the Qwen model, their output token counts are actually much more stable with the Nemotron model.

Despite performance consistently peaking at intermediate context sizes, the scale of the observed degradation appears to be model-dependent and corpus-sensitive. This suggests that while scaling limits exist, they may be influenced by multiple interacting factors rather than representing an absolute ceiling.

\subsection{Corpora Type Sensitivity}

The type of in-context corpus strongly affects both the magnitude of gains and the scaling behavior. We see that the unsupervised corpus scores the lowest by far, and the parallel corpora scores the highest. The relative ranking between corpus types is consistent between both the ChrF++ and COMET score metrics. 

We also observe that the choice of reference language in a parallel corpus produces some direction-specific effects. Although Indonesian is linguistically closer to Javanese and Sundanese compared to English, it does not always guarantee better translation quality. In particular, Indonesian-parallel examples tend to be more helpful for translation from the target language into English, whereas English-parallel examples consistently benefit translation from English into the target language. A possible explanation is that the parallel language helps the model better understand the source language by providing alignment, rather than helping it translate into the target language.

An additional observation is that the instruction-style corpus consistently yields performance comparable to at least one of the parallel settings. Parallel data is often more desirable than monolingual data, as it preserves alignment between the reference and target languages. However, the model is shown to learn almost as well with monolingual instruction-style data. It should be noted that while the total amount of tokens is roughly the same, parallel corpora will always contain roughly half the amount of monolingual target tokens compared to the other monolingual corpora. The number of monolingual target tokens may also play a role in how much information the model gathers through in-context learning. We leave a more comprehensive exploration of the matter to future work.

Finally, different corpora provide different patterns of saturation and degradation as context grows. For example, we can see that the performance disparity between supervised (parallel) and unsupervised (monolingual) corpora is more pronounced in translations from English to the target language, whereas the gap is smaller in the other direction. The effect of scaling the context size is different for different corpora, as seen in Fig.~\ref{fig:comet}. In most cases, unsupervised corpora have the lowest peak and the fastest drop, while others, like instructions and parallel, maintain performance to slightly larger budgets before degrading near the end of the context size range. This suggests that different types of data may be influencing the effectiveness of the model's cognition. Factors such as how information is structured (e.g., aligned pairs vs.\ unaligned text) and how directly it supports the translation mapping can affect both token-efficiency and stability in the many-shot regime.

\section{Conclusion}

We present an empirical study of many-shot ICL up to 1M tokens for low-resource MT, focusing on the Javanese and Sundanese languages. We investigate and characterize how different in-context corpora influence translation performance by systematically scaling the context length from $2^{7}$ to $2^{20}$ tokens and by constructing contexts from three distinct corpus types: monolingual target-language text, instruction-style data, and parallel corpora with English or Indonesian as reference languages. As part of this work, we construct and release synthetic English and Indonesian parallel corpora for Javanese and Sundanese.

Across settings, we found that translation accuracy does not scale linearly with the number of shot tokens. Performance improves initially, but saturates around $2^{14}$ and $2^{16}$ shot tokens, well before maximum context. 
% Moreover, pushing prompts toward the largest context sizes can be counterproductive, with translation quality degrading, and generation becoming less stable near the maximum of the context window size.
Moreover, pushing prompts toward the largest context sizes can be counter-productive in certain configurations, with the degree of degradation depending on model architecture and corpus type. While some settings show substantial quality degradation and generation instability near maximum context (particularly Qwen with certain corpus types), others exhibit a more gradual decline (e.g., Nemotron with parallel corpus for target-to-English translation). These observations suggest that, despite the fact that effective context limits appear clearly, their order of magnitude depends on interactions between model design, corpus characteristics, and task requirements.

These model-specific differences likely stem from distinct architectural design choices: Qwen and Nemotron employ different attention mechanisms and training procedures, though isolating the exact contributing factors would require controlled ablations beyond this study's scope.

These results suggest that the \emph{effective} context for improving MT can be substantially smaller than the maximum context size, motivating careful sampling rather than simply maximizing context usage. While architectural limitations may contribute to degradation, observations suggest that sampling and in-context learning dynamics are also contributing factors. We also observed that the type of in-context corpus influences the translation quality and scaling behavior. Parallel data generally provides the strongest gains, while monolingual text is less effective. A monolingual instruction-style corpus can be surprisingly competitive with parallel corpora, highlighting that formatting and task-following signals may contribute meaningfully to translation. Finally, the relative benefit of English and Indonesian parallel demonstrations depends on translation direction, suggesting that the choice of reference language may interact with context comprehension and target-side generation in non-trivial ways.

\section{Limitation}

This study focused on relatively small decoder-only transformer models, specifically using architectures with around 8B parameters. We have not explored how the same setting would behave under a different model architecture, and we did not systematically vary model size. We focus our experiments on two closely related Austronesian languages, due to compute cost for long-context evaluation and also the inference cost for machine translation. Furthermore, we utilized synthetic data to support our experiments. Although we conducted manual reviews to ensure the translation quality, however, it may not fully resemble human translations.

\section{Future Works}

Future work could further investigate the nuances of utilizing many-shot ICL for MT. A possible area for exploration is developing efficient sampling strategies for a given prompt. Prior studies show that both the selection of in-context examples \cite{zebaze-etal-2025-context} and their specific ordering \cite{zebaze-etal-2025-context, lu-etal-2022-fantastically} significantly influence the translation quality of pretrained language models in few-shot settings.

It is also plausible that larger models may utilize ICL and long-context capabilities more effectively. In that case, many-shot ICL may be an even more competitive alternative for adapting MT systems as finetuning cost scales faster than KV cache resource usage as the model's parameter count grows.

Another promising direction is to mitigate the drop that we observed and discussed in this paper. One's best-case scenario for long in-context learning is that the performance increase is proportional to the additional information provided in context. While this could be an architectural limitation, our experiments show preliminary evidence that some data types or formats can shift the peak or plateau further to the right. 

\bibliography{anthology-used, custom}

\clearpage
\onecolumn
\appendix

\section{Synthetic Parallel Data Construction}
\label{app:parallel}

\begin{figure}[!h]
  \centering
  \setlength{\fboxsep}{8pt}
  \fbox{%
  \begin{minipage}{0.95\columnwidth}
    \ttfamily \footnotesize
    \textbf{Prompt:}\\
    Translate this \{Source Lang Label\} sentence into \{Target Lang Label\}. You must only reply with the translated sentence, no other details are required.\\
    \\
    \{Source Lang Sentence\}
  \end{minipage}
  }
  \caption{Prompt template used to translate the target languages into the reference languages.}
  \label{fig:tlprompttemplate}
\end{figure}

Before translating, we apply a two-stage preprocessing pipeline to select high-quality, diverse samples from the Aya Collection dataset. First, we segment raw text into sentences and apply a filter to remove sentences containing web artifacts (URLs, emails, HTML tags), code syntax, and excessive non-Latin characters. We also enforce constraints on sentence length (5–60 words), character ratios for digits, punctuation, and capitalization, and require proper sentence boundaries with valid starting characters and terminal punctuation.

Next, we employ a diversity-aware sampling strategy that selects sentences to maximize vocabulary coverage while meeting a target token budget of 500K tokens. The sampler prioritizes sentences containing novel or underrepresented words, while filtering out short sentences and those with low average word length. To ensure corpus quality, we perform near-duplicate detection by normalizing text and checking for containment relationships between candidate and selected sentences.

We then utilize OpenAI's GPT5 model accessed through the OpenRouter platform to translate the extracted sentences. The Javanese and Sundanese sentences are translated into English and Indonesian separately. We use the API's default sampling and generation parameters and set GPT5's reasoning mode to minimal.

\section{Random Sampling Results}
\label{app:sampling}

\begin{figure*}
  \includegraphics[width=\columnwidth]{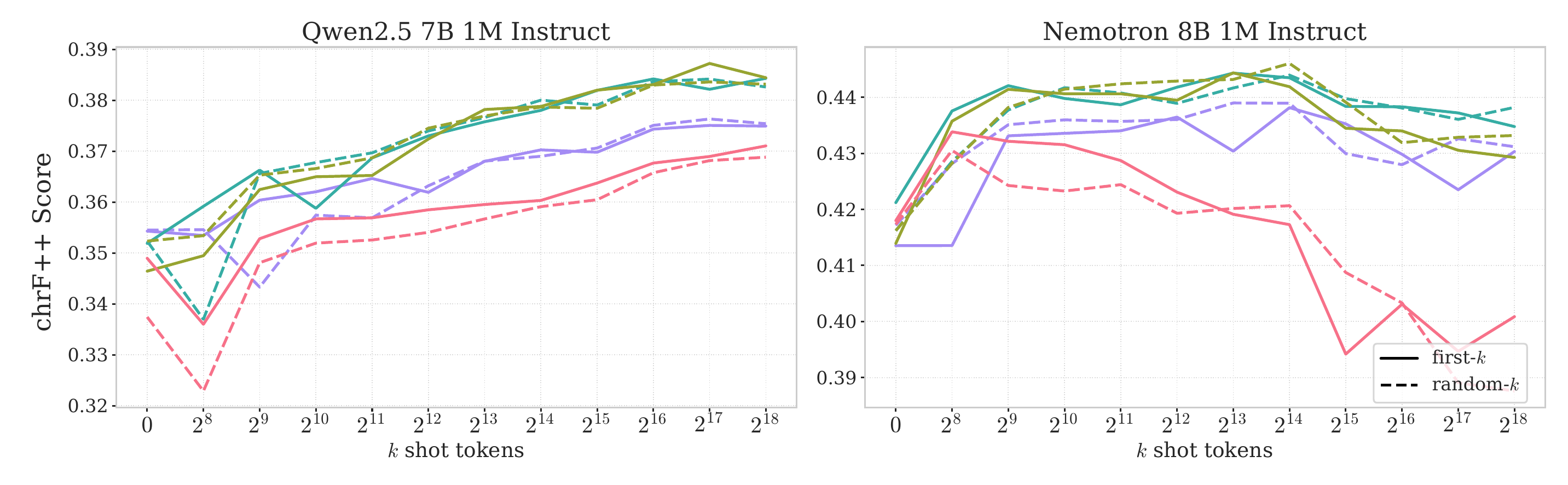}
  \caption{Average ChrF++ score difference between first-$k$ and random-$k$ sampling across all corpus types of Javanese MT experiments, denoted by solid and dotted lines, respectively. Both models show no noticeable performance scaling difference.}
  \label{fig:sampling}
\end{figure*}

To test whether our main scaling trends are sensitive to the construction of our in-context demonstrations, we repeat the evaluation using \textit{random-$k$} sampling, where we randomly sample examples from the same corpus under the same token budget (instead of
always taking the first $k$ examples). Figure~\ref{fig:sampling} shows that the ChrF++ curves under \textit{first-$k$} (solid) and \textit{random-$k$} (dotted) perform similarly for both models across context sizes.

That said, this observation does not rule out more fine-grained positional effects. Random-$k$ sampling changes which examples are included, but it does not explicitly control which demonstrations appear \emph{closest} to the query, where they may have the strongest influence. We leave more tests analyzing the order and positional effects of long-context ICL and MT for future work.

\section{Results Summary}
\label{app:summary_results}

\begin{table}[htbp]
\centering
\caption{chrF++ performance summary across models, corpus types and languages.}
\resizebox{\textwidth}{!}{
\begin{tabular}{llllllll}
\toprule
\multirow{2}{*}{\textbf{Model}} & \multirow{2}{*}{\textbf{Data~Type}} & \multirow{2}{*}{\textbf{Lang}} & \multicolumn{5}{c}{\textbf{Accuracy}} \\
\cmidrule(lr){4-8}
 & & & 0-shot & Best & Best@ & $\Delta$ & $\Delta$\% \\
\midrule
Qwen2.5-7B-Instruct-1M & Unsupervised & JV$\rightarrow$EN & 0.399 & 0.426 & 262K & +0.027 & +6.8\% \\
 & Unsupervised & SU$\rightarrow$EN & 0.406 & 0.436 & 65K & +0.029 & +7.2\% \\
 & Unsupervised & EN$\rightarrow$JV & 0.298 & 0.300 & 262K & +0.002 & +0.8\% \\
 & Unsupervised & EN$\rightarrow$SU & 0.301 & 0.308 & 262K & +0.007 & +2.3\% \\
 & Instructions & JV$\rightarrow$EN & 0.400 & 0.433 & 262K & +0.033 & +8.1\% \\
 & Instructions & SU$\rightarrow$EN & 0.406 & 0.443 & 262K & +0.037 & +9.0\% \\
 & Instructions & EN$\rightarrow$JV & 0.297 & 0.302 & 16K & +0.005 & +1.6\% \\
 & Instructions & EN$\rightarrow$SU & 0.301 & 0.306 & 262K & +0.006 & +2.0\% \\
 & Supervised (EN) & JV$\rightarrow$EN & 0.400 & 0.439 & 262K & +0.039 & +9.9\% \\
 & Supervised (EN) & SU$\rightarrow$EN & 0.406 & 0.447 & 262K & +0.041 & +10.1\% \\
 & Supervised (EN) & EN$\rightarrow$JV & 0.298 & 0.325 & 65K & +0.027 & +9.2\% \\
 & Supervised (EN) & EN$\rightarrow$SU & 0.301 & 0.325 & 131K & +0.024 & +8.0\% \\
 & Supervised (ID) & JV$\rightarrow$EN & 0.401 & 0.441 & 131K & +0.040 & +10.0\% \\
 & Supervised (ID) & SU$\rightarrow$EN & 0.405 & 0.446 & 131K & +0.041 & +10.2\% \\
 & Supervised (ID) & EN$\rightarrow$JV & 0.298 & 0.324 & 131K & +0.026 & +8.7\% \\
 & Supervised (ID) & EN$\rightarrow$SU & 0.301 & 0.322 & 131K & +0.021 & +6.9\% \\
\midrule
Llama-3.1-Nemotron-8B & Unsupervised & JV$\rightarrow$EN & 0.489 & 0.504 & 1K & +0.015 & +3.1\% \\
-UltraLong-1M-Instruct & Unsupervised & SU$\rightarrow$EN & 0.463 & 0.491 & 2K & +0.028 & +6.0\% \\
 & Unsupervised & EN$\rightarrow$JV & 0.359 & 0.375 & 2K & +0.016 & +4.6\% \\
 & Unsupervised & EN$\rightarrow$SU & 0.323 & 0.348 & 256 & +0.024 & +7.5\% \\
 & Instructions & JV$\rightarrow$EN & 0.486 & 0.517 & 16K & +0.031 & +6.3\% \\
 & Instructions & SU$\rightarrow$EN & 0.462 & 0.497 & 16K & +0.035 & +7.7\% \\
 & Instructions & EN$\rightarrow$JV & 0.360 & 0.389 & 16K & +0.029 & +8.1\% \\
 & Instructions & EN$\rightarrow$SU & 0.325 & 0.353 & 8K & +0.028 & +8.7\% \\
 & Supervised (EN) & JV$\rightarrow$EN & 0.487 & 0.517 & 8K & +0.030 & +6.2\% \\
 & Supervised (EN) & SU$\rightarrow$EN & 0.462 & 0.497 & 16K & +0.035 & +7.6\% \\
 & Supervised (EN) & EN$\rightarrow$JV & 0.358 & 0.399 & 32K & +0.041 & +11.5\% \\
 & Supervised (EN) & EN$\rightarrow$SU & 0.323 & 0.358 & 32K & +0.035 & +10.8\% \\
 & Supervised (ID) & JV$\rightarrow$EN & 0.488 & 0.521 & 16K & +0.034 & +6.9\% \\
 & Supervised (ID) & SU$\rightarrow$EN & 0.464 & 0.507 & 16K & +0.043 & +9.3\% \\
 & Supervised (ID) & EN$\rightarrow$JV & 0.359 & 0.395 & 65K & +0.036 & +10.0\% \\
 & Supervised (ID) & EN$\rightarrow$SU & 0.322 & 0.351 & 1K & +0.029 & +9.0\% \\
\bottomrule
\end{tabular}
}
\label{tab:chrfpp_summary}
\end{table}

\begin{table}[htbp]
\centering
\caption{COMET-22 performance summary across models, corpus types and languages.}
\resizebox{\textwidth}{!}{
\begin{tabular}{llllllll}
\toprule
\multirow{2}{*}{\textbf{Model}} & \multirow{2}{*}{\textbf{Data~Type}} & \multirow{2}{*}{\textbf{Lang}} & \multicolumn{5}{c}{\textbf{Accuracy}} \\
\cmidrule(lr){4-8}
& & & 0-shot & Best & Best@ & $\Delta$ & $\Delta$\% \\
\midrule
Qwen2.5-7B-Instruct-1M & Unsupervised & JV$\rightarrow$EN & 0.695 & 0.734 & 262K & +0.039 & +5.6\% \\
 & Unsupervised & SU$\rightarrow$EN & 0.704 & 0.749 & 262K & +0.045 & +6.4\% \\
 & Unsupervised & EN$\rightarrow$JV & 0.723 & 0.728 & 131K & +0.005 & +0.7\% \\
 & Unsupervised & EN$\rightarrow$SU & 0.667 & 0.686 & 262K & +0.020 & +2.9\% \\
 & Instructions & JV$\rightarrow$EN & 0.699 & 0.748 & 262K & +0.050 & +7.1\% \\
 & Instructions & SU$\rightarrow$EN & 0.703 & 0.762 & 262K & +0.058 & +8.3\% \\
 & Instructions & EN$\rightarrow$JV & 0.725 & 0.739 & 262K & +0.014 & +2.0\% \\
 & Instructions & EN$\rightarrow$SU & 0.668 & 0.680 & 262K & +0.012 & +1.8\% \\
 & Supervised (EN) & JV$\rightarrow$EN & 0.696 & 0.752 & 262K & +0.056 & +8.0\% \\
 & Supervised (EN) & SU$\rightarrow$EN & 0.704 & 0.759 & 262K & +0.055 & +7.8\% \\
 & Supervised (EN) & EN$\rightarrow$JV & 0.724 & 0.749 & 262K & +0.025 & +3.4\% \\
 & Supervised (EN) & EN$\rightarrow$SU & 0.668 & 0.714 & 262K & +0.046 & +6.9\% \\
 & Supervised (ID) & JV$\rightarrow$EN & 0.698 & 0.755 & 131K & +0.057 & +8.2\% \\
 & Supervised (ID) & SU$\rightarrow$EN & 0.703 & 0.759 & 262K & +0.056 & +8.0\% \\
 & Supervised (ID) & EN$\rightarrow$JV & 0.726 & 0.740 & 262K & +0.014 & +2.0\% \\
 & Supervised (ID) & EN$\rightarrow$SU & 0.671 & 0.697 & 131K & +0.026 & +3.8\% \\
\midrule
Llama-3.1-Nemotron-8B & Unsupervised & JV$\rightarrow$EN & 0.769 & 0.789 & 512 & +0.021 & +2.7\% \\
-UltraLong-1M-Instruct & Unsupervised & SU$\rightarrow$EN & 0.746 & 0.785 & 2K & +0.039 & +5.2\% \\
 & Unsupervised & EN$\rightarrow$JV & 0.756 & 0.786 & 256 & +0.030 & +4.0\% \\
 & Unsupervised & EN$\rightarrow$SU & 0.712 & 0.744 & 1K & +0.032 & +4.5\% \\
 & Instructions & JV$\rightarrow$EN & 0.764 & 0.799 & 16K & +0.035 & +4.5\% \\
 & Instructions & SU$\rightarrow$EN & 0.745 & 0.785 & 16K & +0.040 & +5.4\% \\
 & Instructions & EN$\rightarrow$JV & 0.759 & 0.787 & 512 & +0.028 & +3.6\% \\
 & Instructions & EN$\rightarrow$SU & 0.711 & 0.744 & 1K & +0.033 & +4.6\% \\
 & Supervised (EN) & JV$\rightarrow$EN & 0.763 & 0.800 & 8K & +0.037 & +4.9\% \\
 & Supervised (EN) & SU$\rightarrow$EN & 0.746 & 0.789 & 8K & +0.043 & +5.8\% \\
 & Supervised (EN) & EN$\rightarrow$JV & 0.756 & 0.797 & 32K & +0.041 & +5.4\% \\
 & Supervised (EN) & EN$\rightarrow$SU & 0.714 & 0.758 & 512 & +0.044 & +6.2\% \\
 & Supervised (ID) & JV$\rightarrow$EN & 0.769 & 0.798 & 8K & +0.029 & +3.8\% \\
 & Supervised (ID) & SU$\rightarrow$EN & 0.751 & 0.796 & 16K & +0.045 & +6.0\% \\
 & Supervised (ID) & EN$\rightarrow$JV & 0.756 & 0.785 & 512 & +0.029 & +3.8\% \\
 & Supervised (ID) & EN$\rightarrow$SU & 0.708 & 0.746 & 512 & +0.038 & +5.4\% \\
\bottomrule
\end{tabular}
}
\label{tab:comet_summary}
\end{table}

\section{Example Prompts}
\label{app:prompts}

\begin{figure}[H]
  \centering
  \setlength{\fboxsep}{8pt}
  \fbox{%
  \begin{minipage}{0.95\columnwidth}
    \ttfamily \footnotesize
    \textbf{Prompt:}\\
    You are Qwen, created by Alibaba Cloud. You are a helpful translation assistant. Your current task is to translate texts as accurately as possible.\\
    \\
\textbf{English}: english

\textbf{Javanese}: javanese

\textbf{English}: The 1965 FIL World Luge Championships on February 6 and 7 in Davos and won the first World Championship title with an advantage of more than a second after four runs, ahead of Petra Tierlich, Ilse Geisler, and Barbara Winter.

\textbf{Javanese}: FIL World Luge Championships 1965 tanggal 6 lan 7 Februari ing Davos lan menang gelar Kejuaraan Dunia sing pertama kanthi kauntungan luwih saka sak detik sawise patang babak, luwih saka Petra Tierlich, Ilse Geisler lan Barbara Winter.

\textbf{English}: Many heads of state liked good personal relations and were unwilling to go to war: Emperor Franz Joseph, who wept as he declared war on Serbia after being misled by the Foreign Minister, and Tsar Nicholas II and Kaiser Wilhelm II, shown as unable to undo their countries’ military mobilization schedules.

\textbf{Javanese}: Akeh kepala negara sing seneng hubungan pribadi sing apik lan ora gelem perang: Kaisar Franz Josef sing nangis ngumumake perang marang Serbia sawise diapusi dening Menteri Luar Negeri, lan Czar Nicholas II lan Kaiser Wilhelm II ditampilake ora bisa ngilangi jadwal mobilisasi militer negarane.

\textbf{English}: Guests and other residents included opera singer Geraldine Farrar, baritone Antonio Scotti, film director and producer D. W. Griffith, novelist F. Scott Fitzgerald, as well as many politicians and diplomats.

\textbf{Javanese}: Tamu lan warga liyane kalebu penyanyi opera Geraldine Farrar, bariton Antonio Scotti, sutradara lan produser film D. W. Griffith, novelis F. Scott Fitzgerald, uga akeh politisi lan diplomat.

\textbf{English}: Four controlled explosions, devices that were somewhat less advanced than the previous attacks, were carried out at Shepherd’s Bush, Warren Street and Oval railway stations, and on a bus in Shoreditch.

\textbf{Javanese}: Papat ledakan sing dikendhaleni, piranti sing luwih kurang maju tinimbang serangan sadurunge, ditindakake ing stasiun sepur Shepherd's Bush, Warren Street lan Oval, lan ing bis ing Shoreditch.

\textbf{English}: Produce a more complex version of this sentence The Nahuatl literature is extensive (possibly the most extensive of all indigenous languages in the Americas), including a relatively large corpus of poetry (see also Nezahualcoyotl).

\textbf{Javanese}: Ngasilake versi sing luwih rumit saka ukara iki Sastra Nahuatl akeh (bisa uga paling akeh saka kabeh basa pribumi ing Amerika), kalebu corpus puisi sing relatif gedhe (deleng uga Nezahualcoyotl).

\textbf{English}: Formulate an answer to this complex question: What is the name of the abolitionist and women’s rights activist identified in The Dinner Party by Judy Chicago?

\textbf{Javanese}: Ngrumusake jawaban kanggo pitakonan sing rumit iki: Apa Jeneng saka ABOLITIONIS AS AS lan aktivis hak-hak wanita sing ditemtokake ing 'The Dinner Party', dening JUDY CHICAGO?

\textbf{English}: Formulate an answer to this complex question: Joseph Wright Harriman was the president of what organization, and the cousin of the future diplomat, statesman, and Governor of New York, W. Averell Harriman?

\textbf{Javanese}: Ngrumusake jawaban kanggo pitakonan sing rumit iki: Joseph Wright Harriman, minangka presiden organisasi apa, lan sepupu diplomat, negarawan lan Gubernur New York ing mangsa ngarep, W. Averell Harriman?

\textbf{English}: The November 24 matchup against the Alabama Crimson Tide was the first game played in the newly completed Legion Field.

\textbf{Javanese}: Generate a more complex version of this sentence Pertandhingan tanggal 24 November nglawan Alabama Crimson Tide minangka pertandingan pertama sing dimainake ing Legion Field sing mentas rampung.\\

    Translate the following sentence from English to Javanese.\\
    English sentence: Dr. Ehud Ur, professor of medicine at Dalhousie University in Halifax, Nova Scotia and chair of the clinical and scientific division of the Canadian Diabetes Association cautioned that the research is still in its early days.\\
  \end{minipage}
  }
  \caption{Example of a prompt used for evaluation on the Qwen2.5-7B-Instruct-1M model with context size of 1024 tokens using the \textbf{parallel English-Javanese} corpus without random sampling for a task to translation from English to Javanese. We remove the part ``You are Qwen, created by Alibaba Cloud.'' for the Nvidia\_Llama-3.1-Nemotron-8B-UltraLong-1M-Instruct model.}
  \label{fig:prompttemplate2}
\end{figure}

\begin{figure}[H]
  \centering
  \setlength{\fboxsep}{8pt}
  \fbox{%
  \begin{minipage}{0.95\columnwidth}
    \ttfamily \footnotesize
    \textbf{Prompt:}\\
    You are a helpful translation assistant. Your current task is to translate texts as accurately as possible.\\
    \\
\textbf{Indonesian}: indonesian

\textbf{Javanese}: javanese

\textbf{Indonesian}: Kejuaraan Dunia Luge FIL 1965 pada tanggal 6 dan 7 Februari di Davos dan memenangkan gelar Kejuaraan Dunia pertamanya dengan keunggulan lebih dari satu detik setelah empat babak, mengungguli Petra Tierlich, Ilse Geisler, dan Barbara Winter.

\textbf{Javanese}: FIL World Luge Championships 1965 tanggal 6 lan 7 Februari ing Davos lan menang gelar Kejuaraan Dunia sing pertama kanthi kauntungan luwih saka sak detik sawise patang babak, luwih saka Petra Tierlich, Ilse Geisler lan Barbara Winter.

\textbf{Indonesian}: Banyak kepala negara yang menyukai hubungan pribadi yang baik dan enggan berperang: Kaisar Franz Josef yang menangis mengumumkan perang terhadap Serbia setelah ditipu oleh Menteri Luar Negeri, dan Czar Nicholas II serta Kaiser Wilhelm II ditampilkan tidak mampu membatalkan jadwal mobilisasi militer negara mereka.

\textbf{Javanese}: Akeh kepala negara sing seneng hubungan pribadi sing apik lan ora gelem perang: Kaisar Franz Josef sing nangis ngumumake perang marang Serbia sawise diapusi dening Menteri Luar Negeri, lan Czar Nicholas II lan Kaiser Wilhelm II ditampilake ora bisa ngilangi jadwal mobilisasi militer negarane.

\textbf{Indonesian}: Tamu dan warga lainnya termasuk penyanyi opera Geraldine Farrar, bariton Antonio Scotti, sutradara dan produser film D. W. Griffith, novelis F. Scott Fitzgerald, serta banyak politisi dan diplomat.

\textbf{Javanese}: Tamu lan warga liyane kalebu penyanyi opera Geraldine Farrar, bariton Antonio Scotti, sutradara lan produser film D. W. Griffith, novelis F. Scott Fitzgerald, uga akeh politisi lan diplomat.

\textbf{Indonesian}: Empat ledakan terkendali, perangkat yang kurang lebih lebih maju daripada serangan sebelumnya, dilakukan di stasiun kereta Shepherd's Bush, Warren Street dan Oval, serta di sebuah bus di Shoreditch.

\textbf{Javanese}: Papat ledakan sing dikendhaleni, piranti sing luwih kurang maju tinimbang serangan sadurunge, ditindakake ing stasiun sepur Shepherd's Bush, Warren Street lan Oval, lan ing bis ing Shoreditch.

\textbf{Indonesian}: Menghasilkan versi yang lebih rumit dari kalimat ini Sastra Nahuatl banyak (mungkin paling banyak dari semua bahasa pribumi di Amerika), termasuk korpus puisi yang relatif besar (lihat juga Nezahualcoyotl).

\textbf{Javanese}: Ngasilake versi sing luwih rumit saka ukara iki Sastra Nahuatl akeh (bisa uga paling akeh saka kabeh basa pribumi ing Amerika), kalebu corpus puisi sing relatif gedhe (deleng uga Nezahualcoyotl).

\textbf{Indonesian}: Merumuskan jawaban untuk pertanyaan yang rumit ini: Siapa nama dari abolisionis dan aktivis hak-hak perempuan yang ditampilkan dalam ‘The Dinner Party’ oleh Judy Chicago?

\textbf{Javanese}: Ngrumusake jawaban kanggo pitakonan sing rumit iki: Apa Jeneng saka ABOLITIONIS AS AS lan aktivis hak-hak wanita sing ditemtokake ing 'The Dinner Party', dening JUDY CHICAGO?

\textbf{Indonesian}: Menyusun jawaban untuk pertanyaan rumit ini: Joseph Wright Harriman, sebagai presiden organisasi apa, dan sepupu dari diplomat, negarawan, dan Gubernur New York di masa depan, W. Averell Harriman?

\textbf{Javanese}: Ngrumusake jawaban kanggo pitakonan sing rumit iki: Joseph Wright Harriman, minangka presiden organisasi apa, lan sepupu diplomat, negarawan lan Gubernur New York ing mangsa ngarep, W. Averell Harriman?\\

    Translate the following sentence from English to Javanese.\\
    English sentence: Dr. Ehud Ur, professor of medicine at Dalhousie University in Halifax, Nova Scotia and chair of the clinical and scientific division of the Canadian Diabetes Association cautioned that the research is still in its early days.\\
  \end{minipage}
  }
  \caption{Example of a prompt used for evaluation on the Qwen2.5-7B-Instruct-1M model with context size of 1024 tokens using the \textbf{parallel Indonesian-Javanese} corpus without random sampling for a task to translation from English to Javanese. We remove the part ``You are Qwen, created by Alibaba Cloud.'' for the Nvidia\_Llama-3.1-Nemotron-8B-UltraLong-1M-Instruct model.}
  \label{fig:prompttemplate3}
\end{figure}

\begin{figure}[H]
  \centering
  \setlength{\fboxsep}{8pt}
  \fbox{%
  \begin{minipage}{0.95\columnwidth}
    \ttfamily \footnotesize
    \textbf{Prompt:}\\
    You are a helpful translation assistant. Your current task is to translate texts as accurately as possible.\\
    \\
Daptar tilu prinsip étika pakait sareng ngumpulkeun data. \\

1. Transparansi sareng kajujuran: Penting pikeun pangumpul data transparan ngeunaan inpormasi anu aranjeunna kumpulkeun, kumaha éta bakal dianggo, sareng sareng saha éta bakal dibagikeun. Pamilon kedah terang ngeunaan tujuan ngumpulkeun data sareng kedah masihan idin anu jelas sareng terang. 2. Hormat pikeun privasi: Inpormasi pribadi anu dibagikeun dina prosés pendataan kedah dijaga rahasia sareng aman. Koléktor data kedah mastikeun yén aranjeunna gaduh ukuran kaamanan anu pas pikeun ngajagaan data pribadi, sareng pamilon kedah gaduh hak aksés, ngarobih atanapi ngahapus datana. 3. Adil jeung non-diskriminasi: Praktek ngumpulkeun data kudu dipandu ku prinsip fairness jeung non-diskriminasi. Ieu ngandung harti yén ngumpulkeun data teu kudu dipaké pikeun ngabentenkeun ka grup tangtu jalma, sarta kolektor data kudu nyokot léngkah pikeun mastikeun yén métode maranéhanana henteu bias jeung teu disproportionately mangaruhan populasi tangtu. Éta ogé kedah ati-ati kalayan kamungkinan konsékuansi tina pembuatan kaputusan sareng profil otomatis.\\

 Ngahasilkeun lima judul unik pikeun novél thriller.\\

 1. "Whispers in the Dark: A Dongeng Suspense" 2. "The Silent Stalker: A Gripping Thriller" 3. "Behind Closed Doors: Secret of the Unknown" 4. "Deadly Intentions: A Chilling Mystery" 5. "Twisted Kaleresan: Petualangan anu Ngeureunkeun Jantung"\\

    Translate the following sentence from English to Sundanese.\\
    English sentence: Dr. Ehud Ur, professor of medicine at Dalhousie University in Halifax, Nova Scotia and chair of the clinical and scientific division of the Canadian Diabetes Association cautioned that the research is still in its early days.\\
  \end{minipage}
  }
  \caption{Example of a prompt used for evaluation on the Qwen2.5-7B-Instruct-1M model with context size of 1024 tokens using the \textbf{instructions corpus} without random sampling for a task to translation from English to Sundanese. We remove the part ``You are Qwen, created by Alibaba Cloud.'' for the Nvidia\_Llama-3.1-Nemotron-8B-UltraLong-1M-Instruct model.}
  \label{fig:prompttemplate4}
\end{figure}

\begin{figure}[H]
  \centering
  \setlength{\fboxsep}{8pt}
  \fbox{%
  \begin{minipage}{0.95\columnwidth}
    \ttfamily \footnotesize
    \textbf{Prompt:}\\
    You are a helpful translation assistant. Your current task is to translate texts as accurately as possible.\\
    \\
"Enya. ali ieu mah sok pipindahan"

Sakali mangsa aya ibu-ibu ti Jakarta anu asup ka dealer Mitsubishi rék mariksakeun mobilna di bengkel.

Hansip : " Saha ngaran manéh ? "

Hansip : " Heueuh kuring mah hayang ngakurkeun jeung jawaban manéh !".

Surah Al Masad atawa Al Lahab nyaéta surat ka-111 dina Al Qur'an. Surat ieu diwangun ku 5 ayat sarta kaasup surat makiyyah. Ngaran Al Lahab dicokot tina kecap Al Lahab anu aya dina ayat katilu anu hartina seuneu nu ngéntab-ngéntab. Poko eusi surat ieu eusina ngeunaan nasib salah saurang pamanna Rosululloh SAW nyaéta Abu Lahab reujeung pamajikanana anu diancam ku siksa naraka.

1 Eusi jeung Tarjamah

Anu dina beuheungna meulit tali tina injuk.

Surat Al Lahab ngisaratkeun yén kamusrikan henteu bisa dipertahankeun sarta moal meunang sanajan jalma-jalma nu ngarojongna tihothat.

Kaca ieu panungtungan diédit 11 Maret 2013, jam 12.20.

Sistem nyéta kumpulan dinu sub sistem atawa bagéan atawa komponén nanaon waé anu mangrupa fisik atawa non fisik anu saling boga hubungan hiji jeung nu sejéna jeung aya kerja sama pikeun ngahontal hiji tujuan nu geus di tangtukeun[1]. Contona awak urang ayeuna miboga sakumpulan bagéan nu tangtu atawa biasa nu di sebut organ, jiga samisalna leungeun, suku, otak, lambung anu di hubungkeun ku aliran darah jeung syaraf anu tungtungna jadi jaringan hirup[1]. Sistem awak ieu boga tujuan nyaéta “Hirup” [1].

Subsistem nyaéta bagéan dinu salah sahiji sistem anu bisa mangrupa fisik atawa non fisik (abstrak)

. Subsistem dijero na boga deui subsistem nu leuwih leutik jeung saterusna anu biasa disebut unsur atawa komponen.[1].

Tujuan sistem mangrupa target atawa sasaran pamungkas anu hayang di hontal ku éta sistem nu aya[1]. Sistem di ciptakeun ku sabab aya tujuana, sistem di jieun ngarah éta tujuan téh teu nyimpang anu akhirna bisa ngurangan résiko kagagalan.[1].

Tilu komponen sistem fungsi atawa subsistem nyaéta Input, proses jeung output[2]. Input nyaéta sagala nanaon waé anu asup kanu jero sistem, prosés nyéta parobahan dinu input tadi jadi output, ari Output nyaéta hasil dinu hiji prosés anu mangrupa tujuan dinu éta ayana sistem[2].

Dinu sistem aya anu disebut lingkungan sistem, lingkungan sistem nyaéta perkara-perkara di luar sistem anu bakal mangaruhan kanu éta ayana sistem[2]. Dua lingkungan sistem nyaéta lingkungan jero (internal) jeung lingkungan luar (eksternal) [1].

Hese nerapkeunana mang. Tong boro milihan jalma nu diblacklist, patugasna teu bisaeun sakadar ngabedakeun jalma nu boga tiket jeung nu henteu oge.

Konfrensi pers teh di mksdkeun supaya dinu latihan sareng pas di mess teu aya sesi wawancara ka essien, kitu kang… \\

    Translate the following sentence from English to Sundanese.\\
    English sentence: Dr. Ehud Ur, professor of medicine at Dalhousie University in Halifax, Nova Scotia and chair of the clinical and scientific division of the Canadian Diabetes Association cautioned that the research is still in its early days.\\
  \end{minipage}
  }
  \caption{Example of a prompt used for evaluation on the Qwen2.5-7B-Instruct-1M model with context size of 1024 tokens using the \textbf{unsupervised corpus} without random sampling for a task to translation from English to Sundanese. We remove the part ``You are Qwen, created by Alibaba Cloud.'' for the Nvidia\_Llama-3.1-Nemotron-8B-UltraLong-1M-Instruct model.}
  \label{fig:prompttemplate5}
\end{figure}

\section{Dataset Sample}
\label{app:data}

\begin{figure}[H]
  \centering
  \setlength{\fboxsep}{8pt}
  \fbox{%
  \begin{minipage}{0.95\columnwidth}
    \ttfamily \footnotesize
    \textbf{Unsupervised sample:}\\
    Ganjarane, ayu lan arume sesami.”
    “Sinau ngarosake lan nyumerepi tunggalipun manungsa, tunggalipun rasa, tunggalipun asal lan maksudipun agesang.”
    “… Ping kalihipun perlu babat lan ngatur papan kangge masang Alif. (Masang Alif punika inggih kedah mawi sarana lampah. Boten kenging kok lajeng dipun canthelaken kemawon, lajeng dipun tilar kados mepe rasukan).”
    “Nulung pepadhane, ora nganggo mikir
    Yen ana isi lumuntur marang sesami.”
    “Nulung tiyang kula tindakaken ing pundi-pundi, sak mangsa-mangsa, sak wanci-wanci.”
    “Ajinipun inggih boten sanes namung aji tekad; ilmunipun ilmu pasrah; rapalipun adilipun Gusti.”
    ” … Suwung pamrih, suwung ajrih, namung madosi barang ingkang sae, sedaya kula sumanggaken dhateng Gusti … “
    “Yen kula ajrih, kenging dipun wastani ngandut pamrih utawi ancas ingkang boten sae.”
    “Luh ingkang medal sangking manah punika, dede luh ipun tangis pamrih, nanging luh peresanipun manah suwung pamrih.”
    ” … Wosipun inggih punika ngupadosi padhang ing peteng; seneng ing sengsara, tunggaling sewu yuta … “
    
    30 gram susu bubuk
    daging sapi gandik 500 gr
    
    Kowe lungoâ€¦ ora pamit aku
    Aku ora nyonoâ€¦ kowe arep lungo
    
    sing luwih patut di contoh yo pak Sudiyo kuwi kang...
    urip mung mampir ngombe lan mung wang sinawang.
  \end{minipage}
  }
  \caption{Example of the crawled Javanese monolingual unsupervised data.}
  \label{fig:jvunspvsample}
\end{figure}

\begin{figure}[H]
  \centering
  \setlength{\fboxsep}{8pt}
  \fbox{%
  \begin{minipage}{0.95\columnwidth}
    \ttfamily \footnotesize
    \textbf{Unsupervised sample:}\\
    Aya oge nu nyebutkeun yen seni Marawis teh nya eta salah-sahiji seni â€œBand tepukâ€ kalawan perkusi sabage alat musik utama. Eusi tina lagu-lagu Marawis biasana ngagambarkeun syair-syair pujian jeung kacintaan manusa ka Nu Nyipta.
    Nu maenkeun seni Marawis kurang leuwih tujuh urang. Unikna baheula mah dina ieu seni sok dimaenkeun ku sadudulur misalna aki, incu, alo, jeung dulur-dulur sejenna. Masing-masing nyepeng hiji alat musik diantarana: hajir, marawis, dumbuk, tamborin, simbar jeung rea rea deui.
    Dina seni marawis aya tilu pola tabeuhan nu beda-beda diantarana: zafin, sarah, zaife (disababaraha daerah aya nu nyebutna zaefah). Zaefin mangrupa pola tabeuhan nu mindeng digunakeun pikeun lagu-lagu pujian ka nabi Muhammad SAW. Wirahmana leuwih kalem teu siga musik karatagan. Sedengkeun pola tabeuhan sarah jeung zaife digunakeun pikeun irama karatagan nu ngagugah rasa sumanget. Tiap daerah miboga pola tabeuhan masing masing kaasup di Cirebon. Ieu hal teh dibalukarkeun akibat ayana proses asimilasi jeung budaya daerah satempat. Nepi ka gerakan tari jeung nyanyianna leuwih variatif. Sedengkeun pola tabeuhan sarah digunakeun pikeun ngarak panganten. Dina seni Marawis, oge dikenal istilah ngepang nu hartina silih bales nabeuh jeung ngangkat. Salian ti acara sunatan jeung kawinan seni marawis sering dipintonkeun dina acara-acara seni budaya Islam.
  \end{minipage}
  }
  \caption{Example of the crawled Sundanese monolingual unsupervised data.}
  \label{fig:suunspvsample}
\end{figure}

\begin{figure}[H]
  \centering
  \setlength{\fboxsep}{8pt}
  \fbox{%
  \begin{minipage}{0.95\columnwidth}
    \ttfamily \footnotesize
    \textbf{Instruction sample:}\\
    \emph{Instruction}:\\
    Nggawe resep kanggo nedha bengi pitik sehat nggunakake úa diwenehi.\\
    \emph{Input}:\\
    Bahan: ayam, brokoli, lombok, bawang putih, jahe\\
    \emph{Output}:\\
    **Ayam Jahe Bawang Putih karo Brokoli lan Mrica Tumis Recipe ** Sajian: 4 Bahan: - 4 dada ayam tanpa balung, tanpa kulit - 2 cangkir kembang brokoli - 1 paprika gedhe, dibiji lan dipotong dadi potongan 1 inci - 3 siung bawang putih , minced - 1 tablespoons seger jahe, grated - Salt lan mrico - 2 tablespoons lenga zaitun Directions: 1. Bumbui dodo pitik karo uyah lan mrico. 2. Kalorake lenga zaitun ing wajan gedhe ing medium-dhuwur panas. Tambahake pitik ing wajan lan masak nganti coklat ing njaba, 6-7 menit. 3. Copot pitik saka wajan lan sisihake. Tambah brokoli, paprika, bawang putih, lan jahe ing wajan. Goreng sayuran nganti 3-4 menit, nganti mulai lembut. 4. Irisan pitik sing wis digodhog dadi potongan-potongan ukuran cokotan lan bali menyang wajan karo sayuran. Nglakoake kabeh bebarengan kanggo liyane 2 menit, nganti pitik wis kebak masak lan sayuran empuk. 5. Sajikan panas lan nikmati nedha bengi pitik sing sehat!

  \end{minipage}
  }
  \caption{Example of the instruction Javanese data, machine translated from the original Alpaca dataset in English by prior work.}
  \label{fig:jvinstsample}
\end{figure}

\begin{figure}[H]
  \centering
  \setlength{\fboxsep}{8pt}
  \fbox{%
  \begin{minipage}{0.95\columnwidth}
    \ttfamily \footnotesize
    \textbf{Instruction sample:}\\
    \emph{Instruction}:\\
    Dina 200 kecap atanapi kirang, nyerat carita ngeunaan beurit nganggo inpormasi ieu:\\

    \emph{Input}:\\
    Beurit cicing di taman hiburan anu ditinggalkeun, sareng éta kedah ngaliwat gerombolan sapi anu ambek pikeun uih deui ka bumina.\\

    \emph{Output}:\\
    Jaman baheula, di taman hiburan kuno anu ditinggalkeun, cicing hiji beurit leutik anu namina Max. Imah Max mangrupikeun sudut anu nyaman handapeun rollercoaster kuno, dimana anjeunna hirup damai sareng sugema, ngurilingan ngumpulkeun tuangeun tuangeun sareng ngawangun sayang na. Hiji poé, Max ngahaja wandered teuing jauh ti imah jeung leungit. Nalika wengi murag anjeunna mendakan dirina di sisi sanés taman, tempat anu teu acan kantos anjeunna angkat. Anjeunna terang anjeunna kedah uih deui ka bumina tapi aya masalah anu ageung: gerombolan sapi ambek ngahalangan jalan. Sapi-sapi éta kabur ti peternakan caket dieu sareng parantos ngarebut taman éta, ngaku salaku wilayahna sorangan. Aranjeunna bakal nagih ka saha waé atanapi naon waé anu wani nyebrang jalanna. Max sieun tapi anjeunna terang anjeunna kedah uih ka bumi. Pikiran pikeun nyéépkeun sapeuting nyalira di daérah anu teu biasa dieusi ku kasieunan. Anjeunna neuteup sabudeureun tur, ningali euweuh jalan sejen, anjeunna ngumpulkeun kawani sarta scurried ka hareup, dodging hooves sarta Ngahindarkeun snorts ambek sapi '. Anjeunna lumpat sagancang-gancangna, nyelap-nyelip sareng ngejat dugi ka tungtungna ngahontal kasalametan pojok sakedik handapeun rollercoaster. Tungtungna aman, anjeunna cuddled up dina sayang na murag saré, exhausted tapi senang mun aya imah.

  \end{minipage}
  }
  \caption{Example of the instruction Sundanese data, machine translated from the original Alpaca dataset in English by prior work.}
  \label{fig:suinstsample}
\end{figure}

\begin{figure}[H]
  \centering
  \setlength{\fboxsep}{8pt}
  \fbox{%
  \begin{minipage}{0.95\columnwidth}
    \ttfamily \footnotesize
    \textbf{Parallel sample:}\\
    \emph{Javanese (Original)}:\\
    Jeff Westbrook, 57, saka Santa Rosa ujar manawa dheweke disalahake dening Wakil Erick Gelhaus sawise ditarik tanggal 21 Agustus ing Cotati, saengga ing sawijining wektu dheweke takon marang Gelhaus: 'Pak, apa ana sing salah karo sampeyan?' 'Aku rumangsa kaya nonton wong sing kudu dakbantu,' ujare Pak Westbrook, manajer program ing perusahaan teknologi informasi, marang CBS saka Gelhaus.\\
    \emph{English (Generated)}:\\
    Jeff Westbrook, 57, from Santa Rosa said he was mistreated by Deputy Erick Gelhaus after being pulled over on August 21 in Cotati, so at one point he asked Gelhaus: “Sir, is there something wrong with you?” “I felt like I was watching someone I needed to help,” Mr. Westbrook, a program manager at an information technology company, told CBS of Gelhaus.\\
    \emph{Indonesian (Generated)}:\\
    Jeff Westbrook, 57, dari Santa Rosa mengatakan bahwa dia disalahkan oleh Wakil Erick Gelhaus setelah dihentikan pada tanggal 21 Agustus di Cotati, sehingga pada suatu saat dia bertanya kepada Gelhaus: 'Pak, apakah ada yang salah dengan Anda?' 'Saya merasa seperti melihat seseorang yang harus saya bantu,' kata Pak Westbrook, manajer program di perusahaan teknologi informasi, kepada CBS tentang Gelhaus.

  \end{minipage}
  }
  \caption{Example of the translated parallel Javanese data, extracted from the Aya collection dataset.}
  \label{fig:jvparallelsample}
\end{figure}

\begin{figure}[H]
  \centering
  \setlength{\fboxsep}{8pt}
  \fbox{%
  \begin{minipage}{0.95\columnwidth}
    \ttfamily \footnotesize
    \textbf{Parallel sample:}\\
    \emph{Sundanese (Original)}:\\
    Rencana aslina nya éta 83 munara Martello, masing-masing dipasang hiji meriam beurat, dina interval sapanjang basisir jeung tilu munara 11 meriam di Sea Houses (Eastbourne), Rye Harbour jeung Dymchurch.\\
    \emph{English (Generated)}:\\
    The original plan was 83 Martello towers, each fitted with a heavy cannon, at intervals along the coast, and three 11-gun towers at Sea Houses (Eastbourne), Rye Harbour, and Dymchurch.\\
    \emph{Indonesian (Generated)}:\\
    Rencana aslinya yaitu 83 menara Martello, masing-masing dipasang satu meriam berat, pada interval sepanjang pesisir dan tiga menara 11 meriam di Sea Houses (Eastbourne), Rye Harbour, dan Dymchurch.

  \end{minipage}
  }
  \caption{Example of the translated parallel Sundanese data, extracted from the Aya collection dataset.}
  \label{fig:suparallelsample}
\end{figure}

\end{document}